\documentclass[lettersize,journal, twoside]{IEEEtran}

\usepackage{amsmath} 
\usepackage{comment}
\usepackage[ruled]{algorithm}
\usepackage{algpseudocode}
\usepackage{paralist}
\usepackage{booktabs}
\usepackage{todonotes}
\usepackage{subfig}
\usepackage{amssymb}
\usepackage{changepage}
\usepackage{microtype}
\usepackage{svg}
\usepackage{siunitx}
\usepackage[noadjust]{cite} 
\usepackage{microtype}
\usepackage{makecell, tabularx}

\setcellgapes{2pt}
\newcolumntype{L}{>{\raggedright\arraybackslash}X}
\usepackage{caption}
\usepackage{changepage}
\usepackage{hyperref}

\begin{document}

\title{
Towards Online Robot Interaction Adaptation to Human Upper-limb Mobility Impairments in Return-to-Work Scenarios
}

\author{Marta Lagomarsino and Francesco Tassi
\thanks{

This work was supported by the Italian National Institute for Insurance against Accidents at Work (INAIL) under ergoCub-CORE project and the Italian Ministry of University and Research (MUR) under the Fondo Italiano per la Scienza (FIS 3), project EPIC (FIS-2024-02654).}
\thanks{The authors are with the Human-Robot Interfaces and Interaction (HRII) Laboratory, Istituto Italiano di Tecnologia, Genoa, Italy.}
\thanks{Corresponding author's email: {\tt\small marta.lagomarsino@iit.it}
}%
\thanks{Digital Object Identifier (DOI): see top of this page.}
}

\markboth{IEEE Robotics and Automation Letters. Preprint Version.}
{Lagomarsino \MakeLowercase{\textit{et al.}}: Towards Online Robot Interaction Adaptation to Human Upper-limb Mobility Impairments}

\maketitle

\begin{abstract} 
Work environments are often inadequate and lack inclusivity for individuals with upper-body disabilities. 
This paper presents a novel online framework for adaptive human-robot interaction (HRI) that accommodates users' arm mobility impairments, ultimately aiming to promote active work participation. Unlike traditional physical HRI approaches that assume able-bodied users, our method integrates a mobility model for specific joint limitations into a hierarchical optimal controller. This allows the robot to generate reactive, mobility-aware behaviour online and guides the user's impaired limb to exploit residual functional mobility. The framework was tested in handover tasks involving different upper-limb mobility impairments (i.e., emulated elbow or shoulder arthritis, and wrist blockage), under both standing and seated configurations with task constraints using a mobile manipulator, and complemented by quantitative and qualitative comparisons with state-of-the-art approaches. Preliminary results indicated that the framework can personalise the interaction to fit within the user's impaired range of motion and encourage joint usage based on the severity of their functional limitations.
\end{abstract}

\vspace{-0.2cm}

\setlength{\textfloatsep}{10pt}
\section{Introduction}

More than one billion people worldwide live with a disability, with nearly 200 million experiencing significant difficulties in daily functioning. 
Although laws support disabled employment, only 27\% of working-age disabled individuals are employed, compared to 56\% of non-disabled individuals \cite{unreportdisability2024}. 
Employed individuals with upper-body limitations are frequently assigned low-value tasks, highlighting persistent labour market barriers. This exclusion reduces independence, social participation, and quality of life, in addition to causing economic losses \cite{socioeconomicdisability}. 
Despite the incredible advances in rehabilitation, including AI-driven physical therapy and robotic-assisted strategies, complete functional restoration is often difficult \cite{proietti2022wearable}. 
With an ageing workforce and rising work-related health conditions, addressing disability employment is more urgent than ever. 

Physically assistive robots offer a valuable opportunity to promote accessibility and independence in the workplace for individuals with upper-body mobility impairments \cite{nanavati2023physically}. Assistive exoskeletons, for example, have shown great potential in compensating for arm muscular weakness and enabling users to perform daily activities \cite{gandolla2021assistive}. However, they also present limitations, including discomfort from additional weight, operational pressures that may cause irritation in long-term usage, and restrictions on natural movement, mobility, and postural balance. 
An alternative approach involves teleoperated robotic systems, where a user remotely controls an external robot or manipulator for assistance in pick-and-place tasks \cite{chen2013robots}. 
While effective in certain scenarios, this method does not exploit residual user upper-limb mobility and may inadvertently contribute to physical inactivity, further reducing motor function over time. 

Beyond wearable and teleoperated robotic assistance, adapting the work environment can enhance functional interactions and improve accessibility for impaired users. 
In this context, 
collaborative robots (CoBots) constitute a promising solution \cite{nanavati2023physically}, as they can online adapt to facilitate interaction. CoBots have already demonstrated their ability to improve ergonomics by optimising the object transfer point \cite{Bestick2018Learning}, 
shared kinodynamics in co-manipulation \cite{tassi2022adaptive}, 
proximity and responsiveness 
\cite{lagomarsino2024pro}.  
However, most existing 
approaches, despite personalising models on physical characteristics such as 
segment lengths, assume that users interacting with robots are fully able-bodied, making robots ineffective for people with mobility limitations \cite{drolshagen2024improving}. The integration of impairments into the robot optimisation problem can promote user comfort and better adapt interactions to individual abilities and needs.

The limited existing studies in human-robot collaboration that address upper-limb impairments focused on fully replacing the functionality of the impaired arm with a robotic system \cite{torielli2024laser, poirier2019voice} %
or minimising hand movement 
\cite{ardon2021affordance}. These approaches overlook the complexity of joint-level limitations. 
Moreover, compensating for human joint impairments by having the robot move more \cite{lagomarsino2025mitigating} or attempting to find an interaction point at the hand pose \cite{meng2022fast} is not always feasible due to task or safety constraints. Most importantly, these strategies risk exacerbating the ``learned non-use'' phenomenon \cite{andre2004functional}, where the impaired arm is gradually ignored, worsening the condition, 
rather than encouraging active job inclusion and autonomy.

\begin{figure}[t]
    \centering
    \includegraphics[width=\linewidth]{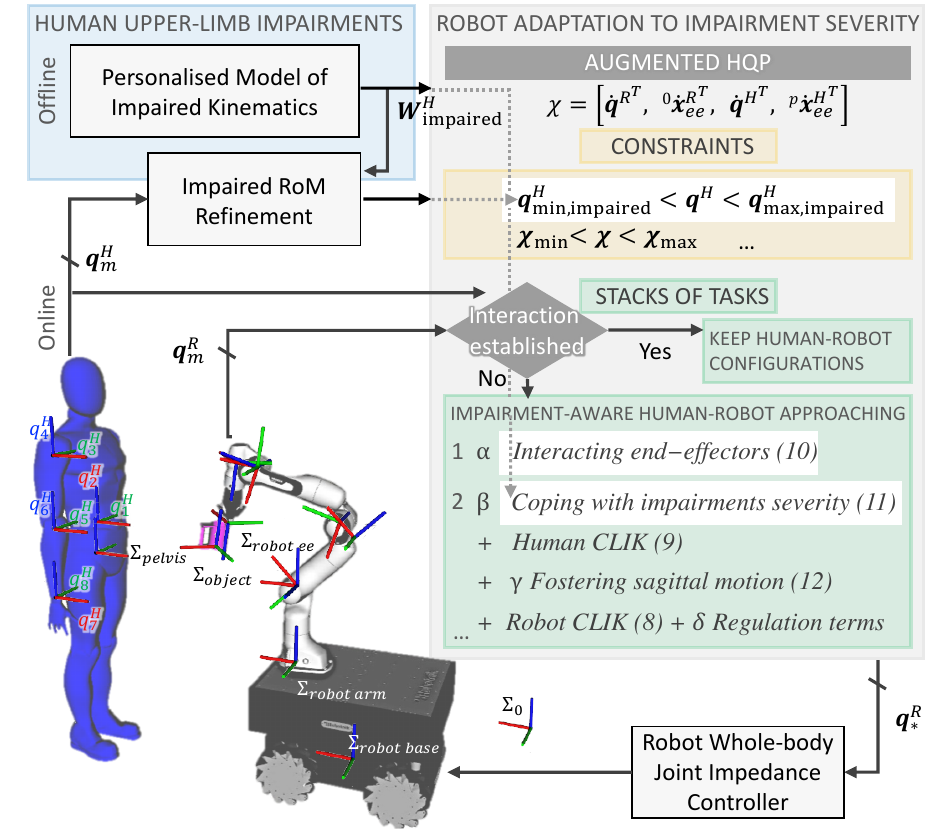}
    \vspace{-0.7cm}
    \caption{\small Overview of the framework to model mobility limitations in impaired users and adapt the HRI strategy based on the severity of specific joints' impairment.}
    \label{fig:schema}
    \vspace{-0.2cm}
\end{figure}

This paper presents the primary yet significant steps towards a framework for online robot adaptation to human upper-limb impairments during interactive tasks, such as robot-to-human handovers or collaborative activities (e.g. gluing, where the human operates the gun and the robot positions the object to be bonded), aiming to promote active work participation by exploiting residual mobility. We propose a subject-specific mobility model that accounts for individual full or partial single-arm joint impairments, body characteristics, and natural redundancies. 
We then formulate an Augmented Hierarchical Quadratic Programming (AHQP) problem 
to simultaneously integrate the impaired-arm mobility model, the user's initial configuration, robot limitations, and specific task requirements as compatible objectives and constraints. 
Since the state space includes the joint and Cartesian velocities of both the human and the mobile manipulator, the optimisation determines not only the final interaction pose but also the Cartesian reference trajectory during the approach phase for both the human and the robot. Additionally, it generates optimal joint-space trajectories that ensure these references are followed while respecting the joint- and task-level constraints of the human-robot system. 
The interaction may begin from any initial user posture (standing or seated) and does not restrict motion of body parts, but once the robot perceives the user's pose at the start of the interaction, we assume a fixed pelvis position, i.e. the user does not walk during the remainder of the task to specifically investigate upper-body compensatory strategies\footnote{Note that the proposed framework could also account for pelvis and whole-body motion, but this work intentionally focus on upper-body compensations.}. 

The main features and contributions of this work are: 
\begin{enumerate}
    \item Our framework models full single-joint and partial multi-joint upper-limb impairments and adapts the human-robot interaction (HRI) position and orientation accordingly, considering the initial human configuration and exploiting residual mobility.  
    \item By treating the human-robot system as a coupled chain in the optimisation, we ensure compliance with joint- and task-level constraints in the interaction approach phase and promote impaired arm movement based on joint severity, helping to counteract ``learned non-use'' and full compensation through the unaffected arm. 
    \item We compare the proposed framework with state-of-the-art approaches through multi-subject handover experiments with emulated impairments, evaluating its performance using quantitative and qualitative metrics. 
\end{enumerate}

\vspace{-0.1cm}
\section{Methods}
\label{sec:methods}
\vspace{-0.1cm}

The proposed framework for online adaptation of robot interaction based on the severity of specific joint impairments is outlined in Fig. \ref{fig:schema}. In the offline phase, we model the user mobility limitations and scale a digital model to automatically match their specific body measurements. In the online phase, we capture the user's initial posture and refine the Range of Motion (RoM) based on modelled residual mobility and current posture. Until the interaction is established, an AHQP problem optimises the robot's approaching strategy according to the user's functional limitations severity, robot constraints, and task requirements, promoting the use of residual mobility.

\subsection{Human Upper-limb Kinematics and Impairment Modelling}
\label{sec:mobility}

In this work, we model human kinematics by assigning a local coordinate frame to each joint, with the x, y, and z-axes defining the axes of rotation for abduction/adduction, flexion/extension, and internal/external rotation, respectively, as 
detailed in \cite{leonori2023bridge}. To capture arm kinematic redundancy, 
we model the arm as a seven Degrees-of-Freedom (DoFs) system, $\boldsymbol{q}^H \!\in\! \mathbb{R}^7$, consisting of two rigid links: the upper arm ($\boldsymbol{l}_\text{humerus}$) 
and the forearm ($\boldsymbol{l}_\text{radius}$). 
Additionally, we assign a frame to the L5 lumbar spine vertebra to model spine flexion with the first joint angle, $q^H_{1}$, and the link, $\boldsymbol{l}_\text{spine}$. 
This results in an overall model with $M=8$ DoFs, where dimension-related parameters can be scaled to accommodate different arm and spine sizes. 
The shoulder is modelled as a spherical joint, including abduction/adduction ($q^H_{2}$), flexion/extension ($q^H_{3}$), and internal/external rotation ($q^H_{4}$). 
The elbow is represented as a hinge-pivot joint, accounting for flexion/extension ($q^H_{5}$) and pronation/supination ($q^H_{6}$) of the forearm. Finally, the wrist joint is modelled as a condyloid joint with flexion/extension ($q^H_{7}$) and ulnar/radial deviation ($q^H_{8}$) of the hand.
Table \ref{tab:dh_params} provides the Denavit-Hartenberg parameters for this kinematic model.

\begin{table}[b!]
    \centering
    \begin{adjustwidth}{0cm}{+3.7cm}
    \caption{\small Denavit-Hartenberg parameters of the adopted human upper-limb kinematic model.}
    \label{tab:dh_params}
    \end{adjustwidth}
    \begin{adjustwidth}{0cm}{-1cm}
    \begin{minipage}[t]{0.52\linewidth}
        \centering
        \footnotesize 
        \begin{tabular}{@{\hspace{0.05cm}}rl@{\hspace{0.2cm}}l@{\hspace{0.1cm}}l@{\hspace{0.2cm}}l@{\hspace{0.05cm}}}
            \toprule
            axis, $i$ & $\boldsymbol{\theta}_{i}$ & $a_i$ & $\alpha_i$ & $d_i$ \\
            \cmidrule[0.4pt](r{0.125em}){1-1}
            \cmidrule[0.4pt](lr{0.125em}){2-5}
            1 & $q_1^H$         & $l_\text{spine}^z$     & --$\frac{\pi}{2}$  & $-l_\text{spine}^y$ \\
            2 & $q_2^H$         & 0                 &  $\frac{\pi}{2}$  & $-l_\text{spine}^x$ \\
            3 & $q_3^H$+$\frac{\pi}{2}$   & 0                 &  $\frac{\pi}{2}$  & 0 \\
            4 & $q_4^H$--$\frac{\pi}{2}$   & -- $l_\text{humerus}^x$  &  $\frac{\pi}{2}$  & -- $l_\text{humerus}^z$ \\
            5 & $q_5^H$+$\pi$     & 0                 &  $\frac{\pi}{2}$  & -- $l_\text{humerus}^y$ \\
            6 & $q_6^H$+$\frac{\pi}{2}$   & -- $l_\text{radius}^y$   &  $\frac{\pi}{2}$  & -- $l_\text{radius}^z$ \\
            7 & $q_7^H$+$\frac{\pi}{2}$   & 0                 &  $\frac{\pi}{2}$  & -- $l_\text{radius}^x$ \\
            hand & $q_8^H$      & 0                 &  0        & 0 \\
            \bottomrule
        \end{tabular}
    \end{minipage}
    \begin{minipage}[h]{0.41\linewidth} 
        \centering
        \vspace{-1.5cm}
        
        \includegraphics[width=\linewidth, trim=0 0 0 0, clip]{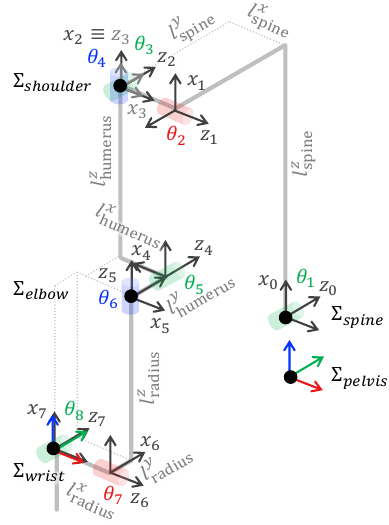}
    \end{minipage}
    \end{adjustwidth}
\vspace{-0.2cm}
\end{table}

Human mobility is traditionally defined by the RoM, i.e., the maximum angular displacement achievable at each joint. An impaired joint has a reduced RoM compared to a healthy one. If $\boldsymbol{q}_\text{min}^{H}$ and $\boldsymbol{q}_\text{max}^{H}$ define the healthy joint limits, the impaired RoM $\big[\boldsymbol{q}_{\text{min}, \text{impaired}}^{H}, \boldsymbol{q}_{\text{max}, \text{impaired}}^{H} \big]$ can be expressed as: 
\begin{align}
    \boldsymbol{q}^H \!\in\! 
    & \big[\boldsymbol{q}_\text{min}^{H} + \Delta\boldsymbol{q}_{\text{min}, \text{impaired}}^{H}, \boldsymbol{q}_\text{max}^{H} - \Delta\boldsymbol{q}_{\text{max}, \text{impaired}}^{H}\big],
\end{align}
where $\Delta\boldsymbol{q}_{\text{min}, \text{impaired}}^{H}$, $\Delta\boldsymbol{q}_{\text{max}, \text{impaired}}^{H}{\small >0}$ are the RoM reductions due to impairments. 
An impaired joint may exhibit varying degrees of reduced mobility depending on the severity of the impairment, which can be estimated through clinical assessments \cite{dshs} or data-driven mobility models \cite{keyvanian2023learning}. Assuming joint-specific impairment indices are available and normalised, we define a diagonal matrix, $\textbf{W}_\text{impaired}^H \!\in\! \mathbb{R}^{M \times M}$, whose diagonal elements encode the severity of impairment for each joint on a scale from $0$ (full functionality) to $1$ (complete impairment)\footnote{Note that the present formulation assumes static and independent RoM representations. However, the proposed framework can accommodate more realistic mobility representations with a configuration-dependent impairment matrix $\textbf{W}_\text{impaired}^H(\boldsymbol{q}^H)$, whose entries vary according to the estimated feasible mobility at the current arm configuration, e.g. using the model in \cite{keyvanian2023learning}.}.  
The resulting RoM reductions are:
\vspace{-0.1cm}
\begin{align} 
\vspace{-3cm}
        & \resizebox{.9\linewidth}{!}{$
		\Delta\boldsymbol{q}_{\text{min}, \text{impaired}}^{H} = \textbf{W}_\text{impaired}^H \big(\min{(\boldsymbol{q}_\text{initial}^{H} - \zeta, \boldsymbol{q}^{H}_m)} - \boldsymbol{q}_\text{min}^H\big), $}  \nonumber\\
        & \resizebox{.9\linewidth}{!}{$
             \Delta\boldsymbol{q}_{\text{max},\text{impaired}}^{H} = \textbf{W}_\text{impaired}^H \big(\boldsymbol{q}_\text{max}^H - \max{(\boldsymbol{q}_\text{initial}^{H} + \zeta, \boldsymbol{q}^{H}_m)}\big),  
         $}
\end{align}
where $\zeta$ 
enables realistic minor joint angle adjustments even in fully impaired joints and potential tracking inaccuracies.
$\boldsymbol{q}_\text{initial}^{H} \!\in\! \mathbb{R}^M$ and $\boldsymbol{q}^H_m \!\in\! \mathbb{R}^M$ are the initial and measured joint positions, enabling RoM reshaping if the user moves beyond the expected limits during the interaction.

\subsection{Online Robot Interaction Adaptation to Human Upper-limb Mobility Impairments}

To adapt the HRI strategy based on the user impairments severity, robot limitations, and specific task requirements, we model the human-robot system as a coupled chain and exploit its redundancy.
To fully utilise this redundancy, we implement a Stack of Tasks (SoT) framework, which prioritises tasks in a strict hierarchical order and allows tasks at the same priority level to be softened. Let $l \!\in\! \mathbb{N}$ represent the total number of hierarchy layers in the SoT, and $k \!\in\! \{1, \ldots, l\}$ denote the levels of priority, with $1$ being the highest and $l$ the lowest. This hierarchy ensures that solutions determined at level $k$ are strictly maintained at the lower priority level $k + 1$, due to the optimality condition defined at each Quadratic Programming (QP) solution \cite{tassi2021augmented}. The general hierarchical problem for a state variable $\boldsymbol{\chi}$ at the $k$-th layer is formulated as:
\vspace{-0.2cm}
\begin{equation}
    \label{eq:qp}
    \min_{\boldsymbol{\chi}} \, \frac{1}{2} \left\| \boldsymbol{A}_k \boldsymbol{\chi} - \boldsymbol{b}_k \right\|^2 \\   \end{equation}
    \begin{equation}
    \text{s.t.} 
    \resizebox{.95\linewidth}{!}{$
    \quad \boldsymbol{C}_1 \boldsymbol{\chi} \leq \boldsymbol{d}_1, \ldots, \boldsymbol{C}_k \boldsymbol{\chi} \leq \boldsymbol{d}_k;
    \quad \boldsymbol{E}_1 \boldsymbol{\chi} = \boldsymbol{f}_1, \ldots, \boldsymbol{E}_k \boldsymbol{\chi} = \boldsymbol{f}_k,
    $} 
    \nonumber 
\end{equation}
where $n_{t_k}$ is the dimension of each $k$-th task, subject to $n_{e_k}$ equality constraints and $n_{i_k}$ inequality constraints. Here, $\boldsymbol{A}_k \!\in\! \mathbb{R}^{n_{t_k} \times s}$, $\boldsymbol{C}_k \!\in\! \mathbb{R}^{n_{i_k} \times s}$, and $\boldsymbol{E}_k \!\in\! \mathbb{R}^{n_{e_k} \times s}$ are matrices, while $\boldsymbol{b}_k \!\in\! \mathbb{R}^{n_{t_k}}$, $\boldsymbol{d}_k \!\in\! \mathbb{R}^{n_{i_k}}$, and $\boldsymbol{f}_k \!\in\! \mathbb{R}^{n_{e_k}}$ are vectors.

\subsubsection{Augmented Hierarchical Quadratic Programming}
In human-robot interactive tasks, rather than having a fixed initial and final target pose, it is advantageous to define a final feasible region and let the optimisation identify the interaction pose and the optimal trajectory to reach it, based on factors such as robot reachability, safe human-robot distance, and human mobility constraints \cite{tassi2021augmented}. Thus, instead of classical Hierarchical Quadratic Programming (HQP), we define an augmented HQP (AHQP) problem \cite{tassi2021augmented}, which does not take the desired human and robot end-effector (ee) trajectories as inputs, but determines them as part of the solution by augmenting the state variable $\boldsymbol{\chi}$ \resizebox{.2\columnwidth}{!}{$  \!\in\! \mathbb{R}^{N + M + 12}$ } as:
\vspace{-0.2cm}
\begin{equation}
    \boldsymbol{\chi} = 
    \begin{bmatrix}
        \dot{\boldsymbol{q}}^{R^T}, 
        {^0\dot{\boldsymbol{x}}}_{ee}^{R^T},
        \dot{\boldsymbol{q}}^{H^T},
        {^p\dot{\boldsymbol{x}}}_{ee}^{H^T} 
    \end{bmatrix}^T,
\end{equation}
where the redundant robotic system is defined by the joint vector $\boldsymbol{q}^R \!\in\! \mathbb{R}^N$, which includes the $N_a$ revolute joints of the manipulator 
and the $N_b$ DoFs of robotic base. 
The latter represent the Cartesian pose of the base $\boldsymbol{x}^R_b = (x, y, \theta)$, comprising the linear and angular positions of the mobile base in the horizontal plane.
${^0\dot{\boldsymbol{x}}}^R_{ee}, {^p\dot{\boldsymbol{x}}}^H_{ee} \!\in\! \mathbb{R}^6$ are the linear and angular Cartesian velocities of the robot ee with respect to the fixed world frame $\Sigma_0$ and of the human hand with respect to the human pelvis frame $\Sigma_\text{pelvis}$, respectively.

In our formulation, once the tasks are defined (see the objective functions in Sec.~\ref{sec:objectives}), the only inputs are the problem constraints, which we can now specify in both human and robot joint-space variables and Cartesian coordinates (see details in Sec.~\ref{sec:constraints}). 
The optimal $\dot{\boldsymbol{q}}^R_*$ obtained from this formulation and $\boldsymbol{q}^R_*$ from its numerical integration are passed to the lower-level joint impedance controller, which generates the necessary compliant actuation torques $\boldsymbol{\tau}^R \!\in\! \mathbb{R}^N$ appropriate for close-proximity interaction as:
\vspace{-0.1cm}
\begin{equation}
    \boldsymbol{\tau}^R = \textbf{K}_q^d (\dot{\boldsymbol{q}}^R_* - \dot{\boldsymbol{q}}^R_m) + \textbf{K}_q^p (\boldsymbol{q}^R_* - \boldsymbol{q}^R_m) + \textbf{g}(\boldsymbol{q}^R_m), 
\vspace{-0.1cm}
\label{eq:commanded_torques}
\end{equation}
where $\dot{\boldsymbol{q}}^R_m, \boldsymbol{q}^R_m \!\in\! \mathbb{R}^N$ are the measured robot joint velocities and positions, $\textbf{K}_q^p, \textbf{K}_q^d \!\in\! \mathbb{R}^{N \times N}$ are the positive definite robot joint stiffness and damping matrices, and $\textbf{g}(\boldsymbol{q}^R_m) \!\in\! \mathbb{R}^N$ is the gravity compensation term. 

\subsubsection{Objective Functions} 
\label{sec:objectives} 

The augmentation of the state variable enables us to define a set of objectives expressed as a function of the state $\bf \chi$, i.e., depend on both the Cartesian coordinates of the ee (or any point along the human-robot kinematic chain) and the joint coordinates. 

\paragraph{Closed-Loop Inverse Kinematics}
For a redundant system, the inverse kinematics problem $\dot{\boldsymbol{q}} = \textbf{J}^{\dagger} \dot{\boldsymbol{x}}$ represents one possible solution 
where $\textbf{J} \!\in\! \mathbb{R}^{6\times N}$ is the Jacobian matrix, 
and $\textbf{J}^{\dagger}$ is its pseudoinverse. This solution can be derived by:
\vspace{-0.2cm}
\begin{equation}
    \min_{\dot{\boldsymbol{q}}} \| \textbf{J} \dot{\boldsymbol{q}} - \dot{\boldsymbol{x}} \|^2, 
\vspace{-0.1cm}
\end{equation}
minimising the error between the desired ee velocity and the one achieved by the joint velocities.
This least-squares formulation allows the use of HQP-based techniques to solve the robot kinematics problem, even in fast and real-time control applications. Furthermore, a Closed-Loop IK (CLIK) scheme \cite{tassi2021augmented} is employed to correct position errors between the desired and actual pose. This modifies the equation to:
\vspace{-0.2cm}
\begin{equation}
    \min_{\dot{\boldsymbol{q}}} \| 
    \textbf{J} \dot{\boldsymbol{q}} - \big(\dot{\boldsymbol{x}} + \textbf{K}_x (\boldsymbol{x} - \boldsymbol{x}_m)\big) \|^2,
\vspace{-0.1cm}
\end{equation}
where $\boldsymbol{x}_m \!\in\! \mathbb{R}^6$ is the measured Cartesian pose, and $\textbf{K}_x \!\in\! \mathbb{R}^{6 \times 6}$ is a positive-definite diagonal gain matrix responsible for ensuring error convergence.
Thus, we can write the optimisation in augmented form, to address the CLIK problem for both the robot:
\vspace{-0.2cm}
\begin{equation}
\label{eq:robot_clik}
    \min_{\boldsymbol{\chi}} 
    \left\| 
    {^0\boldsymbol{J}^R_{ee/0}} \dot{\boldsymbol{q}}^R -
    ({^p\dot{\boldsymbol{x}}}^R_{ee} +
    \textbf{K}^H_x ({^0 \boldsymbol{x}}^R_{ee} - {^0\boldsymbol{x}}^R_{ee, m})) 
    \right\|^2
\end{equation} 
and the human model:
\vspace{-0.2cm}
\begin{equation}
\label{eq:human_clik}
    \min_{\boldsymbol{\chi}} 
    \left\| 
    {^p\boldsymbol{J}^H_{ee/p}} \dot{\boldsymbol{q}}^H -
    \big({^p\dot{\boldsymbol{x}}}^H_{ee} +
    \textbf{K}^R_x ({^p \boldsymbol{x}}^H_{ee} - {^p \boldsymbol{x}}^H_{ee, m})\big) 
    \right\|^2
\end{equation} 
where $\Delta t$ is the control period and ${^0\boldsymbol{x}}^R_{ee, *}{\scriptstyle (t-\Delta t)}$ and ${^p\boldsymbol{x}}^H_{ee, *}{\scriptstyle (t-\Delta t)}$ represent the optimal poses of the ees obtained at the previous time iterations.

\paragraph{Interacting end-effectors}
For interactive tasks, it is crucial that both ees reach the same target pose. Since ${^p\dot{\boldsymbol{x}}}^H_{ee, *}$ and ${^0\dot{\boldsymbol{x}}}^R_{ee, *}$ are now outputs of the AHQP problem, we define a task to minimise the differences between the target poses ${^p\boldsymbol{x}}^H_{ee, *}$ and ${^0\boldsymbol{x}}^R_{ee, *}$ through: 
\begin{align}
\label{eq:interacting_ee_s}
    \min_{\boldsymbol{\chi}} 
    & \left\| 
    {^0\boldsymbol{x}}^R_{ee} \, - \, {^0\boldsymbol{x}}^H_{ee} 
    \right\|^2 = \\
    \min_{\boldsymbol{\chi}} 
    &
    \resizebox{.95\linewidth}{!}{$
    \left\| 
    {^0\boldsymbol{x}}^R_{ee, *}{\scriptstyle(t - \Delta t)} + 
    {^0\dot{\boldsymbol{x}}}^R_{ee} \Delta t \, - \,  
    {^0\boldsymbol{x}}^H_{ee, *}{\scriptstyle(t - \Delta t)} \, -
    \big({^p_0\boldsymbol{R}}_\text{diag} {^p\dot{\boldsymbol{x}}}^H_{ee} + 
    {^0\boldsymbol{r}^H_{p/0}}
    \big)\Delta t
    \right\|^2 \nonumber
    $}
\end{align}
\vspace{-0.5cm}

\noindent where ${^p_0\boldsymbol{R}}_\text{diag}$ is a block diagonal matrix having the rotation matrix between the $\Sigma_\text{world}$ and $\Sigma_\text{pelvis}$ on its diagonal, and ${^0\boldsymbol{r}}^H_{p/0}$ is equal to 0 if the human pelvis is fixed with respect to $\Sigma_\text{world}$. 
It is worth noting that it may be necessary to include an offset between the two target poses, e.g. reflecting the corresponding grasping points for the handover object.

\paragraph{Coping with impairments severity} 
To counteract the ``learned non-use'' phenomenon, 
the residual mobility and joint functionality of each individual should be fully exploited. We therefore define a task that tunes the usage of human arm DoFs based on the joint impairment severity, minimising the velocity (and thus exertion) of significantly impaired joints while engaging healthier ones:
\vspace{-0.1cm}
\begin{equation}
    \min_{\boldsymbol{\chi}} \| \textbf{W}_\text{impaired}^H \dot{\boldsymbol{q}}^H \|^2, 
\label{eq:constr_arm}
\vspace{-0.2cm}
\end{equation}
where $\textbf{W}_\text{impaired}^H$ is the matrix defined in Sec.~\ref{sec:mobility}. 

\paragraph{Fostering sagittal motion}
A low-priority task promotes human motion within the sagittal plane, encouraging more natural intermediate human configurations during the human-robot approach phase \cite{cavanaugh1999kinematic} without interfering with higher-priority tasks. 
This task initially penalises frontal-plane motion and is gradually deactivated as the distance $d_\text{sagittal}^R$ between the robot ee and the desired object grasping point decreases. We implement this by pre-multiplying each human joint velocity 
by a diagonal matrix $\textbf{S}^H_\text{non-sagittal}$:
\begin{equation}
    \min_{\boldsymbol{\chi}} \| \textbf{S}_\text{non-sagittal}^H{\scriptstyle (t)} \dot{\boldsymbol{q}}^H \|^2
\label{eq:foster_saggital}
\vspace{-0.2cm}
\end{equation}
with each diagonal element $s_i = s_i(d_\text{sagittal}^R{\scriptstyle (t)})$ of $\textbf{S}_\text{non-sagittal}$:
\vspace{-0.1cm}
\begin{equation}
    \label{eq:attention}
    \vspace{0.1cm}
    \small 
    \!s_i
     =
    \begin{cases}
        1, \vspace{-0.05cm}
            & \!\!\! \mbox{\small if \normalsize} (\boldsymbol{h}_i \cdot \boldsymbol{n}_\text{sagittal} \sim 0\quad \wedge \\ & {\quad d_\text{sagittal}^R \geq d_\text{max}}) \\
        \!\frac{1}{2}\! \Bigl[ 
        1\!-\!\cos{\!\Bigl(
        \dfrac{d_\text{sagittal}^R - d_\text{min}}
        {d_{\text{max}} - d_{\text{min}}}
        \pi\Bigr)} \Bigr], \vspace{-0.15cm}
            & \!\!\! \mbox{\small if \normalsize} (\boldsymbol{h}_i \cdot \boldsymbol{n}_\text{sagittal} \sim 0 \quad \wedge \\ & {\quad d_\text{min}\! <\! d_\text{sagittal}^R}\! \leq\! d_{\text{max}}) \\
        0, 
            & \!\!\! \mbox{\small otherwise.} \normalsize
    \vspace{-0.4cm}
    \end{cases}
    \normalsize
\end{equation}
The goal is to minimise the displacement of the $i$-th human joint if its rotation axis $\boldsymbol{h}_i$ is orthogonal to the normal vector $\boldsymbol{n}_\text{sagittal}$ of the sagittal plane, i.e. the inner product between $\boldsymbol{h}_i$ and $\boldsymbol{n}_\text{sagittal}$ is close to zero. The parameters $d_\text{min}$ and $d_\text{max}$ define the range over which this task is smoothly deactivated. 

\paragraph{Regularisation terms}
The regularisation term $\min_{\boldsymbol{\chi}} \| \dot{\boldsymbol{q}}^R \|^2$ produces minimum-norm stable solutions by keeping the joint velocities as small as possible.

\subsubsection{Constraints}
\label{sec:constraints} 
We define joint-space and task-space constraints to regulate the AHQP optimisation of the human-robot system within a shared workspace. The joint-space constraints are specified separately for the human and robot, focusing on their respective RoM and joint velocity limits: 
\vspace{-0.2cm}
\begin{gather}
    \boldsymbol{q}_{\text{min}} \leq \boldsymbol{q}{\scriptstyle(t - \Delta t)} + \dot{\boldsymbol{q}}{\scriptstyle(t)}\Delta t \leq \boldsymbol{q}_{\text{max}}, \nonumber
    \\
    \dot{\boldsymbol{q}}_{\text{min}} \leq \dot{\boldsymbol{q}}{\scriptstyle(t)} \leq \dot{\boldsymbol{q}}_{\text{max}}. 
\vspace{-0.5cm}
\end{gather}
For human joints $\boldsymbol{q}^H$, the constraints account for mobility limitations due to impairments (Sec.~\ref{sec:mobility}). For robot joints $\boldsymbol{q}^R$, the constraints consider the physical and actuator limits. 
We define a set of task-space constraints to establish a feasible region for the ee poses and velocities of both the human and robot, specifically for ${^0\boldsymbol{x}}^R_{ee}$, ${^p\boldsymbol{x}}^H_{ee}$, ${^0\dot{\boldsymbol{x}}}^R_{ee}$, and ${^p\dot{\boldsymbol{x}}}^H_{ee}$: 
\vspace{-0.2cm}
\begin{gather}
    \boldsymbol{x}_{\text{min}} \leq \boldsymbol{x}{\scriptstyle(t - \Delta t)} + \dot{\boldsymbol{x}}{\scriptstyle(t)}\Delta t \leq \boldsymbol{x}_{\text{max}}, \nonumber
    \\
    \dot{\boldsymbol{x}}_{\text{min}} \leq \dot{\boldsymbol{x}}{\scriptstyle(t)} \leq \dot{\boldsymbol{x}}_{\text{max}}, 
\vspace{-0.5cm}
\end{gather}
where $\boldsymbol{x}_{\text{min}}$ and $\boldsymbol{x}_{\text{max}}$ denote the Cartesian poses bounds, and $\dot{\boldsymbol{x}}_{\text{min}}$ and $\dot{\boldsymbol{x}}_{\text{max}}$ the linear and angular velocity limits. 
These constraints can be written in an augmented form as a function of $\boldsymbol{\chi}$, both at position
and velocity levels. 

\begin{algorithm}[b]
\caption{\small Online Interaction Adaptation to Impairments}
\label{alg:handover}
\begin{algorithmic}[1]
\State Define severity of human joint impairments: $\boldsymbol{W}^H_\text{impaired}$
\State Initial set of inequality constraints: $\boldsymbol{C}_1$, $\boldsymbol{d}_1$
\State $k \gets 2$
\While{performing interactive task}
    \State Measure $\boldsymbol{q}^R_m, \dot{\boldsymbol{q}}^R_m$, ${^0\boldsymbol{x}}^R_{ee, m}$, ${^p\boldsymbol{x}}^H_{ee, m}$
    \State Refine impaired RoM
    \newline
    \textcolor{gray}{\Comment{\emph{Stack of tasks definition / update}} 
    } 
    \If{interaction established} 
        \State {Keep human and robot configurations}
    \Else 
    \textcolor{gray}{\Comment{\emph{\textit{Approach / Reconfiguration} phase}}
    }
        \State $\boldsymbol{A}_1, \boldsymbol{b}_1 \gets \alpha$ \text{Interacting end-effectors \eqref{eq:interacting_ee_s}}
        \State $\boldsymbol{A}_2, \boldsymbol{b}_2 \gets {\small \beta}$ {\text{Coping with impairments severity \eqref{eq:constr_arm}}  
        \begin{flushright}
        + \text{Human CLIK \eqref{eq:human_clik}} 
        + $\gamma$ \text{Fostering sagittal motion \eqref{eq:foster_saggital}}
        \end{flushright}
        \begin{flushright}
        + \text{Robot CLIK \eqref{eq:robot_clik}}
        + $\delta$ \text{Regularisation terms}
        \end{flushright}}
    \EndIf
    \State Update 
    constraints with online data
    \State $\boldsymbol{\chi}_* \gets$ Solve the QP problem \eqref{eq:qp}
    \State $\boldsymbol{\tau}^R \gets$ Send robot torques commands \eqref{eq:commanded_torques}
\EndWhile
\end{algorithmic}
\end{algorithm}

\subsubsection{Task Priorities in Interactive Tasks}
The proposed AHQP problem aims to adapt the approach and configuration of human-robot interactive tasks (e.g. handovers, drilling, or polishing) to the severity of user-specific mobility impairments.
To this end, we define a SoT including joint-space constraints to enforce impaired RoM and robot limits, and Cartesian safety/task constraints (Sec.~\ref{sec:constraints}). 
The highest-priority task guides the human and robot ees towards interaction in Cartesian space (Sec.~\ref{sec:objectives}, Eq.~\ref{eq:interacting_ee_s}). 
At the second priority level, we set the task that minimises the displacement of impaired joints based on their functional limitation severity (Eq.~\ref{eq:constr_arm}), thereby promoting conscious use of residual mobility. 
Additionally, when the sagittal-plane distance between ees is large, the auxiliary task to encourage sagittal movement is activated (Eq.~\ref{eq:foster_saggital}) that slightly softens the previous objective and, with the human CLIK task, generates a more natural optimal human configuration $\boldsymbol{q}^H_*$. 
Finally, robot-related components of $\boldsymbol{\chi}$ are regulated through the robot CLIK task and its regularisation term. 

The interaction is considered established when: 
\begin{inparaenum}[(i)]
    \item the error between the desired and actual robot ee poses, ensuring sufficient tracking accuracy, and
    \item the relative human-robot ee pose error fall below a threshold $\epsilon$.
\end{inparaenum}
Once these conditions are met, the algorithm transitions to maintaining the interaction configuration by minimising $\boldsymbol{\chi}$ (Alg. \ref{alg:handover}).

\section{Experiments and Results}
\label{sec:experiments_results}

To demonstrate the key features of our framework, we conducted simulations and a multi-subject experiments in handover tasks (see video at \href{https://youtu.be/8FfaMXwcMns}{https://youtu.be/8FfaMXwcMns}). First, simulations modelled full and partial upper-limb impairments to demonstrate the robot's ability to adapt interaction poses across different impairments, human configurations, and transferred objects. Second, we verified that the optimisation of the human-robot coupled chain satisfies joint-level and task constraints during the pre-handover approaching phase and promotes residual mobility use based on impairment severity. Third, we compared a multi-subject experiment with emulated impairments against state-of-the-art approaches. 

\begin{figure*}[!t]
    \centering
    \vspace{-0.5cm}
    \includegraphics[width=\linewidth]{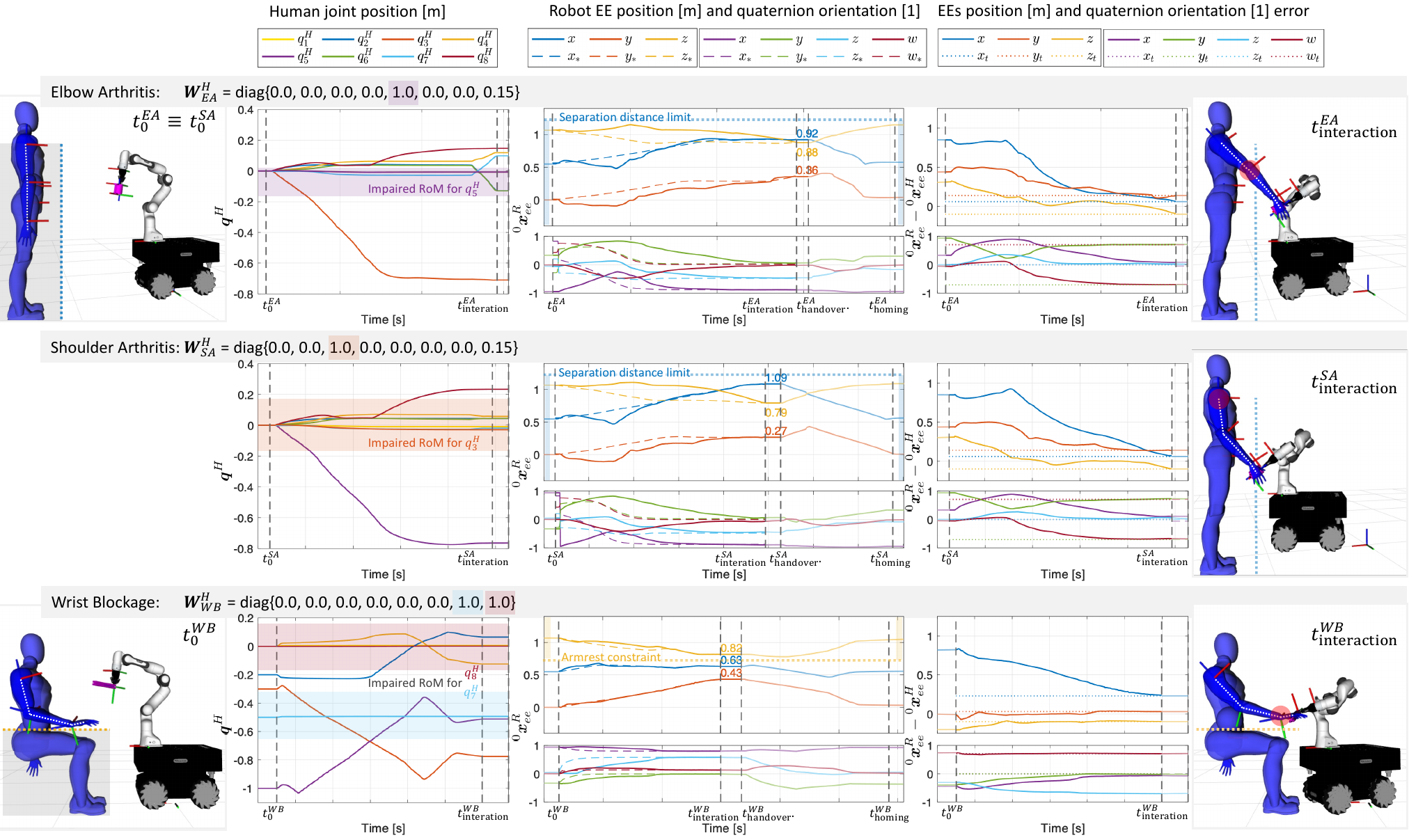}
    \vspace{-0.7cm}
    \caption{\small Online adaptation results for various complete single-joint impairments, showing RViz visualisation of initial and final configurations.
    From left to right, the plots depict optimised human joint angles, promoting residual mobility while respecting impaired RoM; robot ee desired (dashed) vs. actual (solid) positions and orientations, highlighting accurate tracking and compliance with task constraints; and the relative error between human and robot end-effectors, confirming successful establishment of interaction with different objects.
    \vspace{-0.5cm}}
    \label{fig:handover_sim1}
\end{figure*}

\subsection{Experimental Setup and Protocol}

Experiments involved transferring an object from the MOCA mobile manipulator \cite{wu2019teleoperation} to the human in three phases: reciprocal 
approach, physical 
exchange, and post-handover retreat. 
MOCA's $7$-DoF robotic arm and $3$-DoF mobile base provide $4$ degrees of redundancy, which, combined with $2$ degrees of redundancy in the human kinematic model, yield $6$ degrees of redundancy for the coupled human-robot system. 
The AHQP was solved using ALGLIB on an onboard Intel Core i7 processor (8 cores, 8 GB RAM) running Ubuntu 20.04. 

Ten healthy participants (seven male, three female; age $27.8 \!\pm \!2.0$ years), naive to the study purpose, were recruited from the students and research personnel of Istituto Italiano di Tecnologia
under Protocol IIT\_HRII\_ERGOLEAN 156/2020, approved by the ethics committee ASL Genovese N.3. 
Heights ranged from $1.58$ \si{m} to $1.93$ \si{m}, deliberately chosen to evaluate adaptation across anthropometric characteristics. 

We employed the MVN Biomech suit (Xsens Tech.BV) with inertial measurement unit sensors to measure body segment lengths and record the user's kinematics. 
A URDF (Universal Robot Description Format) model of the human kinematic chain (Sec.~\ref{sec:mobility}) was defined, specifying the corresponding joints and links. Healthy RoMs for the model joints were derived from ergonomic tables \cite{booher1994nasa} and used as reference limits for the corresponding DoFs of the adopted kinematic representation and axis convention.
For the links, we used mannequin meshes from \cite{leonori2023bridge}, which we scaled to each participant's measured body segment lengths during real-world experiments. This allowed us to create a digital twin of the human body—modelled as a robot—for running the optimisation and visualising the optimal joint configurations suggested by the AHQP.  

Human joint velocity limits, $\boldsymbol{\dot{q}}_\text{min}^{H}$ and $\boldsymbol{\dot{q}}_\text{max}^{H}$, were set empirically to $\pm 2.5$ \si{rad}/\si{s} and $\zeta = 0.17$ \si{rad} $\sim$10 \si{deg}. Task space limits were a separation distance $d_\text{s} = 0.25$ \si{m} from the human pelvis or feet and set wide for the other components ($\pm 5$ \si{m} in position and $\pm \pi$ \si{rad} in orientation), and linear and angular velocity limits of $\pm 10$ \si{m}/\si{s} and $\pm \pi$ \si{rad}/\si{s}. In the seated scenario, we imposed a chair armrest constraint preventing both the human arm and the robot-grasped object from lowering the $z$-coordinate below $0.70$\si{m} in $\Sigma_0$. 
Manipulator limits followed manufacturer specs. 
The Pinocchio library  
was used for 
human and robot kinematics and their Jacobians. 
Control gains were defined as: $\small \textbf{K}^R_x = \text{diag}\{10,10,10,2,2,2\}$, $\small \textbf{K}^H_x = \text{diag}\{40,40,40,40,40,40\}$, $\small \textbf{K}^p_q = \text{diag}\{1,1,1,10,10,10,10,2,2,2\}$, $\small \textbf{K}^d_q = \text{diag}\{0.1,0.1, \newline 0.1,1,1,1,1,0.2,0.2,0.2\}$ and were chosen to achieve accurate joint-space tracking during motion, while maintaining safe interaction levels in case of collisions.
Parameters $d_\text{min}$ and $d_\text{max}$ were set to $0.1$ and $0.2$ \si{m}.
The AHQP optimisation directly provided optimal joint angles for RViz visualisation.

The task weights were $\alpha,\beta = 100$ for the most relevant tasks at the first and second priority levels, $\gamma = 10$, and $\delta = 0.001$ for the lowest-priority regularisation term, included to ensure smooth trajectories. It should be noted that the relative proportions of task weights matter rather than their absolute values. 
Upon interaction establishment (position/orientation error $ < \epsilon \!=\! 0.01$ \si{m} and $10$ \si{deg}), 
the robot released the object and returned to a homing configuration $\boldsymbol{q}^R_\text{homing}$ through a joint-space postural task $\min_{\boldsymbol{\chi}} \| \boldsymbol{q}^R \!-\! \boldsymbol{q}^R_\text{homing} \|^2$.
A task minimising pelvis angle displacement was included, allowing trunk bending only when strictly needed.

\vspace{-0.1cm}
\subsection{Method Validation} 
\label{sec:simulations}
\vspace{-0.1cm}

Simulations were conducted to demonstrate the method ability to adapt the HRI position and orientation under different upper-limb impairments, initial human configurations, and object types. The framework was tested on both complete single-joint and partial multi-joint impairments.

\subsubsection*{Complete single-joint impairments}
We modelled three common arm impairments via $\boldsymbol{W}_\text{impaired}^H$ (reported in Fig. \ref{fig:handover_sim1}): \textit{Elbow Arthritis} (EA), restricting elbow flexion $q^H_5$; \textit{Shoulder Arthritis} (SA), restricting shoulder flexion $q^H_3$;
and \textit{Wrist Blockage} (WB), often affecting prosthesis or wheelchair users, limiting wrist flexion $q^H_7$ and radial/ulnar deviation $q^H_8$.
A factor of $0.15$ was empirically assigned to the last element of $\boldsymbol{W}_\text{impaired}^H$ to reflect the naturally lower priority of wrist flexion 
compared to shoulder and elbow motion.
Two scenarios were considered: a standing user receiving a mug and a seated user receiving an aluminium profile. Human and robot grasp points were predefined based on object affordances. 

Fig. \ref{fig:handover_sim1} shows the optimisation results for adapting the handover process based on the receiver's complete single-joint mobility limitations. 
The first two rows of the figure show snapshots from the RViz interface, illustrating how the framework, starting from the same initial human and robot configurations in the standing condition (left, at $t_0^{\scriptscriptstyle EA/SA}$), adapts the final interaction position and orientation (right, at $t_\text{interaction}^{\scriptscriptstyle EA/SA}$) to accommodate the modelled impairment.
The bottom row of the figure depicts the handover scenario for a seated human receiving the profile, with the RViz snapshots on the left illustrating the initial condition ($t_0^{\scriptscriptstyle WB}$) and the right frame showing the established interaction configuration at $t_\text{interaction}^{\scriptscriptstyle WB}$. From left to right, the plots show the optimised human joint angles, 
robot ee desired (dashed) vs. actual (solid) positions and orientations (in quaternion form), 
and the relative error between human and robot ees. 
In the first column, the shaded areas indicate the refined impaired RoM boundaries due to the modelled joint impairment. Notably, the optimised human joint angles remained within these adjusted limits during the approach and interaction phases while encouraging movement in healthy joints. Task constraints were respected: in the standing condition, the separation distance $d_\text{s}$ was ensured, and in the seated condition, the human arm and the robot did not cross the $z$-coordinate boundary representing the armrest. 
The central plots demonstrate that the robot adhered to these constraints and smoothly returned to the homing pose after the handover. 
The robot tracked the reference pose generated by the optimal controller, achieving the handover pose with mean position and orientation errors of $0.009$ \si{m} and $0.14$ \si{rad}, respectively. 
The final column of plots shows that interaction was successfully established for different handover objects, with the final offset at convergence corresponding to the transformation between the respective object grasping points.

\subsubsection*{Partial multi-joint impairments}

\begin{figure}[b]
    \centering
    \begin{adjustwidth}{-0.5cm}{-0.6cm}
    \includegraphics[width=\linewidth]{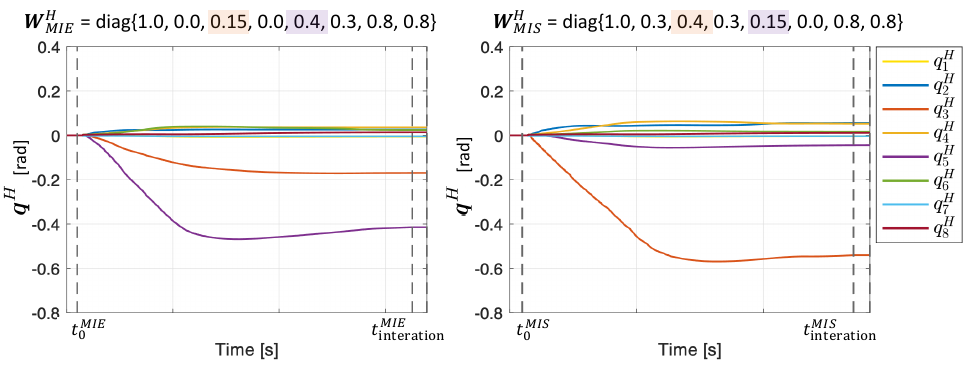}
    \vspace{-.7cm}
    \end{adjustwidth}
    \caption{\small Human redundancy resolution advised by AHQP optimisation for two conditions of partial multi-joint mobility impairments: elbow-dominant (left) and shoulder-dominant (right). 
    \vspace{-0.1cm}
    }
    \label{fig:handover_sim2}
\end{figure}

Two partial multi-joint impairment conditions were also tested to showcase that our method adapts the interaction and resolves human redundancy based on the severity of functional limitations. 
We modelled  
impairments across the sagittal arm joints with different restriction levels at the shoulder, elbow, and wrist. Here, “partial” denotes conditions that moderately reduce RoM but may cause different discomfort levels during use. 
Fig. \ref{fig:handover_sim2} shows that the algorithm resolved redundancy to adapt to these limitations, even when the residual RoM remained relatively wide. 
In the multi-impairment condition with a more severely affected elbow (MIE), joint limits were $q_{3,\text{impaired}}^{\scriptscriptstyle MIE} \!=\! {\small [-2.7,1.8]}$, $q_{5,\text{impaired}}^{\scriptscriptstyle MIE} \!=\! {\small [-1.9,0.02]}$, and $q_{8,\text{impaired}}^{\scriptscriptstyle MIE} \!=\! {\small [-0.2,0.2]}$ \si{rad}. 
In the shoulder-dominant condition (MIS), limits were $q_{3,\text{impaired}}^{\scriptscriptstyle MIS} \!=\! {\small [-1.9,1.3]}$, $q_{5,\text{impaired}}^{\scriptscriptstyle MIS} \!=\! {\small [-2.7,0.02]}$, and $q_{8,\text{impaired}}^{\scriptscriptstyle MIS} \!=\! {\small [-0.2, 0.2]}$ \si{rad}. 
This mobility-aware redundancy resolution in the hierarchical optimisation led to distinct robot compensations. 
The average deviation between the robot 
trajectories in MIS and MIE conditions was $0.16$ \si{m} in position and $1.35$ \si{rad} in orientation, evidencing personalised handover behaviour.

\vspace{-0.1cm}

\subsection{Real-world Experiments and Baseline Comparisons} 
\label{sec:exp}
We conducted multi-subject experiments to evaluate the impact of embedding a user-specific impairment model within the robot optimisation loop. To the best of our knowledge, prior approaches compute the object transfer point offline without modelling joint-level impairments. Existing methods typically adapt the interaction to the user's segment lengths and either (i) minimise hand displacement or (ii) select configurations assumed comfortable for unimpaired users, disregarding heterogeneous mobility profiles. Functional mobility embeddings have been explored in \cite{liu2025grace}, but only to identify feasible interaction regions rather than to optimise the interaction point and approach trajectory. 
We therefore compared our method against two representative impairment-agnostic baselines in the most informative settings, i.e., those in which joint-level impairments substantially affect the interaction strategy. For both baselines, the transfer point was computed offline without considering impairments, and the robot executed the handover at the resulting location.

\subsubsection*{Metrics and statistics}
Statistical analyses were performed on objective kinematics metrics and questionnaire data. Normality was assessed using the Anderson–Darling test and paired t-test or Wilcoxon signed-rank tests were applied accordingly.
Objective metrics included:

(i) Mean compensation of functioning joints~\cite{lagomarsino2025mitigating}
\vspace{-0.2cm}
$$
\small
\bar{\Psi} \!=\! \frac{1}{K_f} \sum\nolimits_{k=1}^{K_f}
\left\| (\mathbf I - \mathbf W^H_{\text{impaired}})
\big(\boldsymbol q^H[k] - \boldsymbol q^H_{\text n} \big) \right\|^2,
\vspace{-0.2cm}
$$
computed separately for arm and trunk ($\bar{\Psi}_{\text{arm}}$ and $\bar{\Psi}_{\text{trunk}}$) w.r.t. a nominal configuration $\boldsymbol{q}^H_{\text n}$ associated with comfortable manipulation in unimpaired users~\cite{pheasant2018bodyspace}, defined by $90^{\circ}$ elbow flexion and neutral alignment of the remaining joints. 

(ii) Motion jerkiness
\vspace{-0.6cm}
$$ \hspace{1.8cm}
\small
J \!=\! \Delta t \sum\nolimits_{k=1}^{K_f} \|\dddot{\boldsymbol q}^H[k]\|^2,
\vspace{-0.2cm}
$$
where $\Delta t$ is the Xsens sampling period and $\dddot{\boldsymbol q}^H \!\in\! \mathbb{R}^{M+2}$ is the jerk vector obtained by numerically differentiating joint velocities. Lower values of $J$ correspond to smoother motion.

\subsubsection*{Comparison with nominal configuration}
\begin{figure}[t]
    \centering
    \includegraphics[width=\linewidth]
    {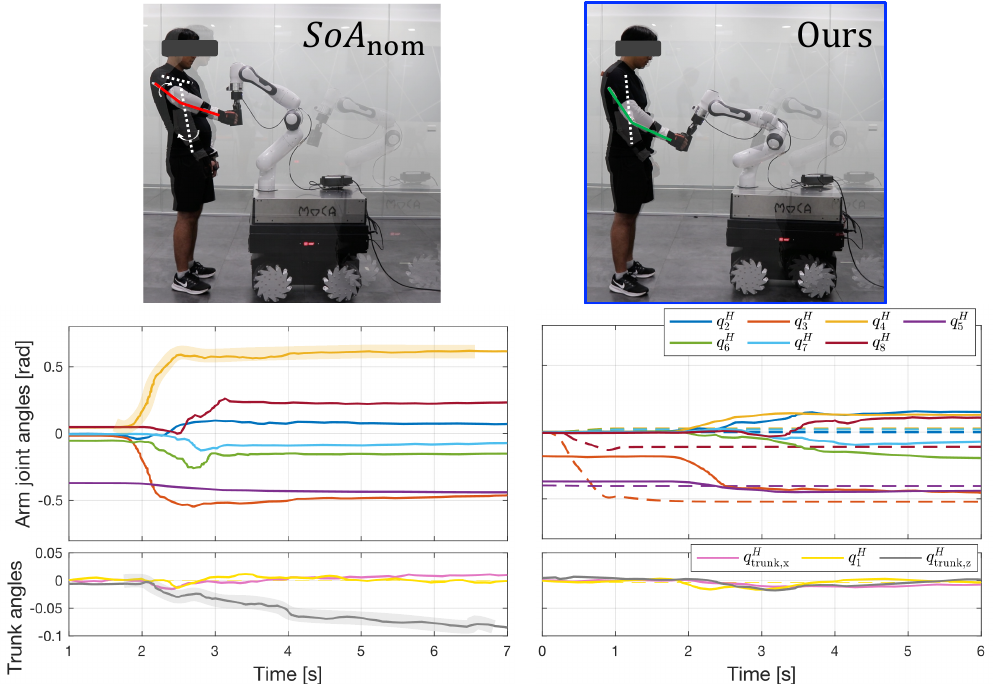}
    \vspace{-0.4cm}
    \caption{\small Standing scenario with emulated elbow flexion impairment (EA): comparison of an impairment-agnostic nominal transfer-point baseline (${\textit{SoA}}_\text{nom}$) and the proposed mobility-aware handover.}
    \label{fig:standing_EA}
    \vspace{-0.2cm}
\end{figure}

The first comparison considered the nominal configuration $\boldsymbol{q}^H_{\text n}$ commonly associated with comfortable manipulation in unimpaired users~\cite{pheasant2018bodyspace}. In the literature, interaction points are often selected around mid-range joint configurations based on standard RoM tables~\cite{booher1994nasa, Bestick2018Learning}, assuming (consistent with ergonomic tools such as Rapid Upper Limb Assessment) that postures near RoM limits increase physical effort. This approach neglects individual mobility differences.  
The comparison with this baseline, denoted ${SoA}_\text{nom}$, was conducted in the standing scenario under an emulated EA impairment, simulated by attaching two rigid links to the upper and forearm and locking the elbow flexion. Each participant performed three robot-to-human handovers per condition. 
Fig.~\ref{fig:standing_EA} reports results for a participant. With the elbow impairment emulated, the ${\mathrm{SoA}}_{\text{nom}}$ handover induced marked compensations: the subject rotated the trunk (decrease in $q^H_{\text{trunk},z}$; grey line in the bottom-left plot) and elevated and retracted the shoulder.
In addition, we observed increased shoulder internal rotation ($q^H_4$, dark yellow line) and forearm pronation ($q^H_6$, green line), resulting in a biomechanically unnatural posture. In contrast, our mobility-aware strategy exploited residual mobility to achieve a more 
natural interaction. The robot facilitation guided the user towards a posture aligned with the optimised configuration $\boldsymbol{q}^H_*$ (dashed lines in Fig.~\ref{fig:standing_EA}-right), albeit with a slightly different redundancy resolution. A small temporal delay relative to the optimal human trajectory was observed in some participants, as they could not visualise the optimised posture 
over time.

\begin{figure}[t]
    \centering
    \vspace{-0.2cm}
    \includegraphics[width=0.95\linewidth]{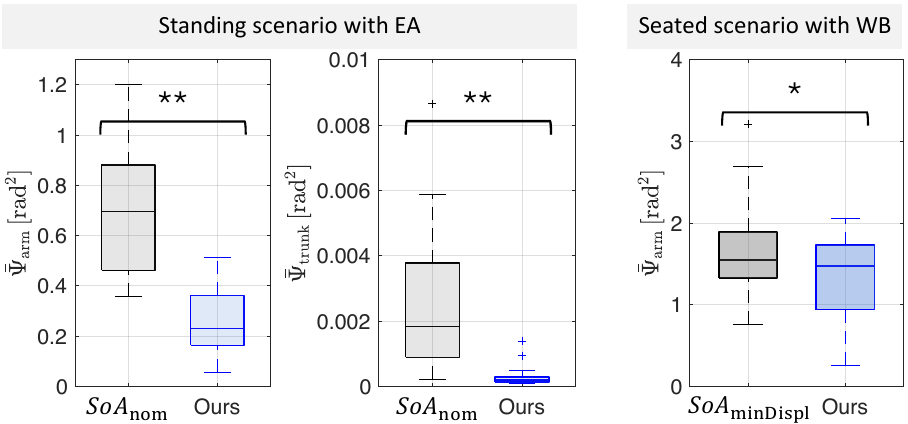}
    \vspace{-0.2cm}
    \caption{\small Statistical comparison of mean arm and trunk compensation costs between baselines (black) and proposed mobility-aware approach (blue) in standing and seated scenarios (**$p\!<\!0.01$,*$<\!0.05$).}
    \label{fig:statistical_analysis}
    \vspace{-0.2cm}
\end{figure}

Across participants, ${\textit{SoA}}_\text{nom}$ induced heterogeneous compensatory patterns, including trunk rotation, extension, and lateral flexion; shoulder elevation, abduction, and internal rotation; often accompanied by forearm pronation and wrist exertion. 
Our method significantly reduced $\bar{\Psi}_\text{arm}$ and $\bar{\Psi}_\text{trunk}$ (see box plots in Fig.~\ref{fig:statistical_analysis}-left) and arm and trunk $J$ ($p\!<\!0.01$). These results indicate that our strategy reduced compensatory effort and induced smoother, more natural approach motion.

\subsubsection*{Comparison with minimum hand-displacement} 
\begin{figure}[t]
    \centering
    \includegraphics[width=\linewidth]
    {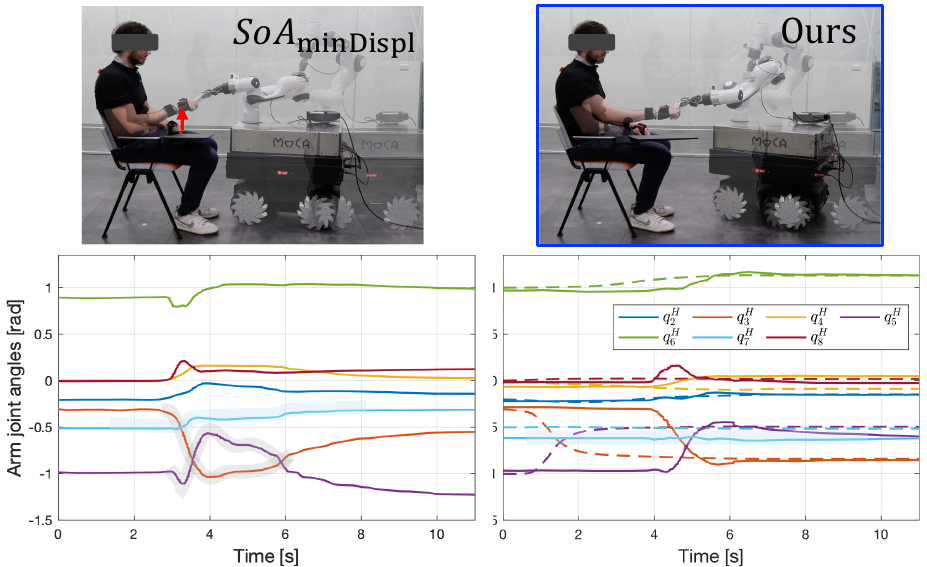}
    \vspace{-0.6cm}
    \caption{\small Seated scenario with emulated wrist blockage (WB): comparison of the minimum-hand-displacement baseline (${\textit{SoA}}_\text{minDispl}$) and the proposed mobility-aware handover.}
    \label{fig:sitted_WB}
    \vspace{-0.3cm}
\end{figure}

The second comparison used a minimum hand-displacement baseline~\cite{ardon2021affordance}: the closest feasible Cartesian transfer point from the initial hand position was computed under the task constraint of avoiding collision with the arm rest. This baseline, denoted ${\text{SoA}}_\text{minDisp}$, was evaluated in the seated condition with an emulated WB impairment, simulated by attaching a rigid link blocking wrist movement. Each condition was repeated three times. 
Fig.~\ref{fig:sitted_WB} reports results for one participant. Under the armrest constraint, ${\text{SoA}}_\text{minDisp}$ selected a handover at roughly the same horizontal distance but higher elevation. The participant advanced and then retracted the arm, yielding overshoot in elbow ($q^H_5$, purple line in left plot) and shoulder flexion ($q^H_3$, orange), indicating a misprediction of the robot's final pose. Small wrist flexion ($q^H_7$, light blue) was also observed despite the physical constraint. By contrast, our wrist-blockage-aware optimisation resulted in a slight arm extension with no wrist displacement (right plot).
Across participants, ${\text{SoA}}_\text{minDisp}$ induced, especially in taller subjects with longer upper- and forearm segments, significantly larger peak elbow/shoulder flexion and wrist displacement ($p\!<\!0.01$). Overall, our method led to a significantly lower mean compensation of the arm functioning joints (Fig~\ref{fig:statistical_analysis}-right).

\begin{figure}[t]
    \centering
    \includegraphics[width=1.07\linewidth]{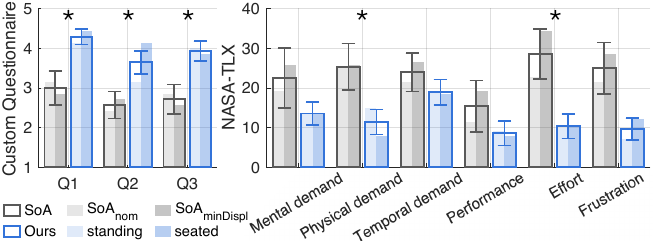}
    \vspace{-0.6cm}
    \caption{\small Results of subjective questionnaires (*$p<0.05$).}
    \label{fig:questionnaires}
    \vspace{-0.2cm}
\end{figure}

\subsubsection*{Questionnaires}
Participants' self-reports 
favoured the proposed mobility-aware optimisation over baselines (see Fig. \ref{fig:questionnaires}). On the 5-point Likert-scale custom questionnaire (1 = strongly disagree, 5 = strongly agree), participants reported greater ability to adopt a comfortable posture (Q1, $p\!=\!0.01$), a reduced need for compensatory movements (Q2, $p\!=\!0.02$), and increased confidence/safety during interaction (Q3, $p\!=\!0.03$). On the NASA-TLX, \textit{physical demand} and \textit{effort} were significantly lower with the proposed method ($p\!=\!0.02$ and $p\!=\!0.05$); the remaining scales trended positively but did not reach statistical significance.

\vspace{-0.4cm}
\section{Discussion and Conclusions}
\label{sec:discussions_conclusions}
\vspace{-0.1cm}

We presented a novel online method for adapting HRI to cope with individual upper limb impairments. 
The latter were modelled and integrated into a hierarchical optimal controller that, instead of pre-defining the interaction pose, optimises the relative approaching by exploiting human residual mobility. 

Across standing and seated handovers with emulated impairments, state-of-the-art baselines (impairment-agnostic nominal transfer point and minimum hand displacement) led to larger compensations in functioning joints, highlighting the need for individual mobility profiles for inclusive HRI. In contrast, the proposed method consistently reduced the mean compensation cost for arm and trunk, 
improved motion smoothness, 
and kept the human configuration within the impaired range of motion while adapting to impairment severity and task constraints. 
Despite these encouraging results, we observed a temporal delay and minor deviations in users' redundancy resolution. Future work will investigate user guidance for configuring functioning joints (e.g., visual or vibrotactile feedback), close the loop with human-tracking corrections when posture drifts from the advised configuration, and explore interaction-force detection at the end-effector to detect established interaction. Moreover, we will extend the framework to more complex interactive tasks involving users with real motor impairments and markerless motion capture.  

\addtolength{\textheight}{-6cm}   






\small

\bibliographystyle{IEEEtran}

\bibliography{bibliography}

\end{document}